\pdfoutput=1

\documentclass[11pt]{article}

\usepackage[final]{acl}

\usepackage{times}
\usepackage{latexsym}
\usepackage{graphicx}
\usepackage{floatrow}
\usepackage{wrapfig}
\usepackage{amsmath}
\usepackage{amssymb}
\usepackage{listings}
\usepackage{tikz}
\usepackage{booktabs}
\usepackage{array}
\usepackage{makecell}
\usepackage{multirow}
\usepackage{soul}
\usepackage{xcolor}
\usepackage[most]{tcolorbox}
\newfloatcommand{capbtabbox}{table}[][\FBwidth]
\usepackage[T1]{fontenc}
\newcommand{\eg}{e.g.,\ }
\newcommand{\ie}{i.e.,\ }

\definecolor{highlightgreen}{RGB}{204,255,102}
\newcommand{\hlgreen}[1]{{\sethlcolor{highlightgreen}\hl{#1}}}

\definecolor{highlightblue}{RGB}{103,255,204}

\usepackage[utf8]{inputenc}

\usepackage{microtype}

\usepackage{inconsolata}

\title{Parrot: Enhancing Multi-Turn Instruction Following for \\ Large Language Models}

\newcommand*\samethanks[1][\value{footnote}]{\footnotemark[#1]}

\author{Yuchong Sun$^1$\thanks{Equal contributon. This work was performed when Yuchong Sun was visiting Kuaishou as an intern.}, Che Liu$^2$\samethanks[1], Kun Zhou$^3$\samethanks[1], Jinwen Huang$^2$, Ruihua Song$^1$\thanks{Corresponding author.}, \\ \textbf{Wayne Xin Zhao$^1$\samethanks[2], Fuzheng Zhang$^2$, Di Zhang$^2$, Kun Gai$^2$}  \\
$^1$Gaoling School of Artificial Intelligence, Renmin University of China\\
$^2$Kuaishou
$^3$School of Information, Renmin University of China\\
\texttt{\{ycsun,rsong\}@ruc.edu.cn}, \texttt{liuche03@kuaishou.com} \\
}

\begin{document}
\maketitle
\begin{tikzpicture}[remember picture,overlay,shift={(current page.north west)}]
\node[anchor=north west,xshift=3.2cm,yshift=-2.5cm]{\scalebox{0.5}{\includegraphics[width=2cm]{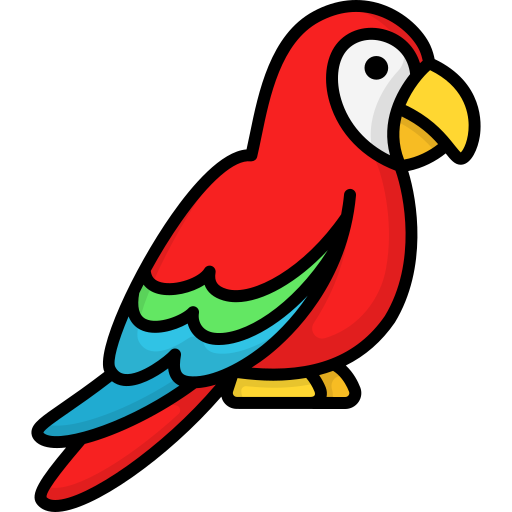}}};
\end{tikzpicture}
\begin{abstract}

Humans often interact with large language models (LLMs) in multi-turn interaction to obtain desired answers or more information.
However, most existing studies overlook the multi-turn instruction following ability of LLMs, in terms of training dataset, training method, and evaluation benchmark.
In this paper, we introduce \textbf{Parrot}, a solution aiming to enhance multi-turn instruction following for LLMs.
First, we introduce an efficient but effective method for collecting multi-turn instructions that feature human-like queries, such as anaphora and ellipsis.
Second, we propose a context-aware preference optimization strategy to further enhance LLMs for complex queries in multi-turn interaction.
Moreover, to quantitatively evaluate LLMs in multi-turn instruction following, we manually build a multi-turn benchmark derived from existing ones.
Extensive experiments show that Parrot improves current LLMs by up to \textbf{7.2\%} in multi-turn instruction following.
Our dataset and codes will be open-sourced to facilitate future research~\footnote{https://github.com/kwai/KwaiYii/Parrot}.

\end{abstract}

\section{Introduction}

Large language models (LLMs)~\cite{openai2022chatgpt, OpenAI2023GPT4TR, alpaca, vicuna,xu2023baize,ding2023ultrachat,li2023humpback,zhou2023lima,zhao2023llmsurvey} have demonstrated their strong capability in understanding a range of human instructions.
By leveraging synthetic or human-created instructions to fine-tune the LLaMA model~\cite{touvron2023llama,touvron2023llama2}, a series of studies~\cite{alpaca, vicuna,xu2023baize,ding2023ultrachat,li2023humpback,zhou2023lima} achieve promising results, even in some benchmarks~\cite{vicuna,alpaca_eval} performing close to ChatGPT and GPT-4.
However, most of these studies and benchmarks overlook the multi-turn instruction following ability of LLMs, which is a more common demand in real-world scenarios.

\begin{figure}[t]
    \centering
    \includegraphics[width=\linewidth]{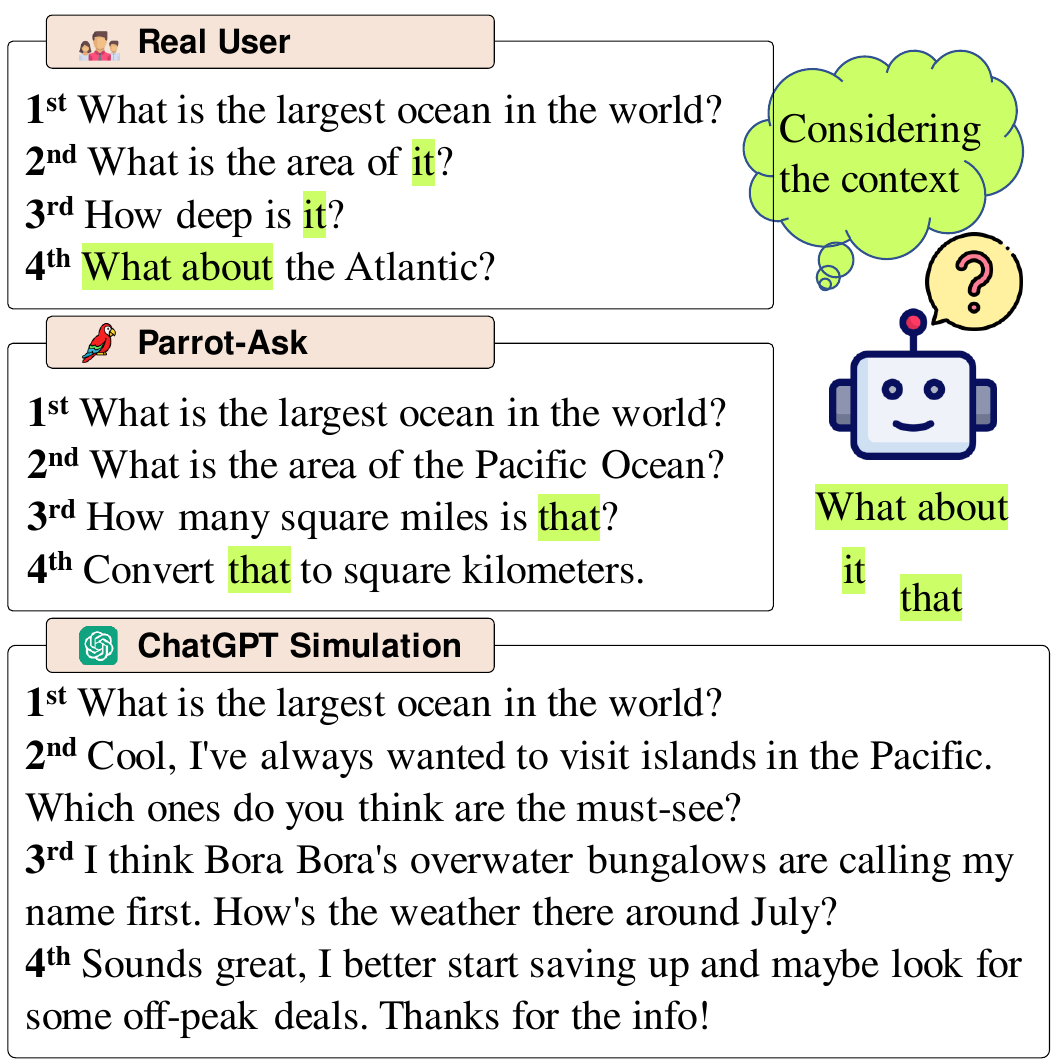}
    \caption{
    In multi-turn interactions, user queries often require LLMs to effectively utilize contextual information, \eg \hlgreen{anaphora} and \hlgreen{ellipsis}. Directly using ChatGPT to simulate users can not fully mimic the above real-world occasions, while our Parrot-Ask trained on real-world conversations can better human-like queries.}
    \label{fig:teaser}
\end{figure}

Developing LLMs capable of multi-turn interaction presents significantly greater challenges compared to single-turn interaction. 
In terms of collecting instruction tuning data, as the number of turns increases, the cost of manual annotation for data collection rises considerably, making it economically unaffordable. 
Existing user logs with ChatGPT, such as ShareGPT, are limited in quantity, and approximately 60\% of the data does not exceed three turns, which is still not enough to achieve satisfactory performance~\cite{vicuna}.
Some LLM-based multi-turn instruction generation methods, even after employing specific prompts to let ChatGPT act as a user~\cite{ding2023ultrachat}, usually produce information-complete queries that lack common features found in human queries, such as anaphoras and ellipses, as shown in Fig.~\ref{fig:teaser}.
In terms of training strategies, most of the current models depend on straightforward supervised fine-tuning (SFT) methods~\cite{vicuna,ding2023ultrachat,xu2023baize}, without specific optimization design for complex queries involving anaphora and ellipsis in multi-turn interactions. 
It may mislead the LLM to neglect or hallucinate the context when generating responses.

To overcome the above challenges, we propose Parrot, which aims to facilitate the development of LLMs with stronger multi-turn instruction following capabilities.
First, we introduce an efficient but effective approach for automatically collecting multi-turn instruction tuning data.
Instead of designing complicated prompts to make powerful LLMs (such as ChatGPT or GPT-4) act as users and generate queries, we propose training a Parrot-Ask model based on a smaller LLM to learn features found in human queries from a small number of real user dialogues, and then it can be used for generating human-like queries.
Second, to enhance the capability of LLMs in handling complex queries during multi-turn interactions, we propose a Context-aware Preference Optimization (CaPO) strategy. We first select queries that heavily rely on contextual information to obtain accurate responses, and then construct negative responses by simulating common error scenarios such as ignoring context or misunderstanding anaphora. Subsequently, we leverage these pairs to optimize the LLM's preferences so that they can better exploit contextual information when generating responses. 

Moreover, for a quantitative assessment of LLMs' multi-turn abilities, we ask annotators to expand the MT-Bench benchmark~\cite{zheng2023llmjudging} that originally consists of two-turn queries, to an eight-turn MT-Bench++ benchmark. MT-Bench++ includes complex queries like anaphoras, ellipses, and topic transitions, better reflecting real multi-turn interaction.
We conduct extensive experiments on MT-Bench and MT-Bench++.
Compared with previous approaches, our proposed method can generate higher-quality multi-turn instruction tuning data with more human-like queries.
Our multi-turn instruction dataset combined with our proposed CaPO strategy improves the current LLMs by 7.2\% in multi-turn instruction following evaluation.

We summarize our contributions as follows:
\begin{itemize}
    \item We propose Parrot, a novel method with a new instruction dataset for enhancing the multi-turn instruction following capability of LLMs.

    \item We design CaPO, a training strategy that simulates common errors in multi-turn conversation, and learns to avoid them in generation.

    \item Our model trained on Parrot dataset with CaPO achieves superior performance among 13B open-source LLMs, especially for the multi-turn instruction following capabilities.
\end{itemize}
\section{Related Work}

\subsection{Instruction Tuning for LLMs}
Instruction tuning plays an important role in inspiring the instruction following ability of LLMs and aligning with humans~\cite{wang2022supernli,wei2021finetuned,ouyang2022instructgpt,openai2022chatgpt,OpenAI2023GPT4TR}. Due to the expensive costs to collect human-annotated instruction tuning data~\cite{DatabricksBlog2023DollyV2,ouyang2022instructgpt}, recent works explore leveraging the powerful LLMs to generate instruction-response pairs in an automatic manner~\cite{alpaca,ding2023ultrachat,xu2023baize,wang2022self,xu2023wizardlm,peng2023instructiongpt4,gpt4all}.
Self-Instruct~\cite{wang2022self} designs seed prompts as examples to prompt GPT-3~\cite{brown2020gpt3} to generate instructions.
Alpaca~\cite{alpaca} adopts the same pipeline to collect instruction-response pairs using ChatGPT and then fine-tune a LLaMA model~\cite{touvron2023llama}.
Humpback~\cite{li2023humpback} proposes instruction back-translation that trains an LLM to generate instructions for web corpus. However, they mainly focus on single-turn instructions. 
Baize~\cite{xu2023baize} collects multi-turn instructions by leveraging ChatGPT to generate dialogues in a self-chat manner. 
UltraChat~\cite{ding2023ultrachat} utilizes two ChatGPT APIs to play the roles of user and assistant respectively.
Vicuna~\cite{vicuna} adopts user-ChatGPT logs from the ShareGPT platform for instruction tuning.
A concurrent work also attempts to train a user simulator to collect instruction tuning data~\cite{kong2023platolm}.
However, these multi-turn instruction data still have several drawbacks, such as less detailed responses, not human-like instruction, or a limited number of turns.

\subsection{Evaluation of Instruction Following}
The current benchmarks for LLMs mainly focus on single-turn evaluation~\cite{hendrycks2020mmlu,zhong2023agieval,srivastava2022bbh,alpaca_eval,he2023cello,zhou2023instruction-following-evaluation}. 
MMLU~\cite{hendrycks2020mmlu} and Big-bench~\cite{srivastava2022bbh} are designed as multiple-choice questions to measure the knowledge and reasoning ability of LLMs. 
AGIEval~\cite{zhong2023agieval} constructs human-centric evaluation for LLMs from standardized exams.
However, the above kinds of evaluation obey the nature of the open-ended generation of LLMs and cannot reflect the ability to follow user instructions~\cite{zheng2023llmjudging}.
Alpaca-Eval~\cite{alpaca_eval} builds a single-turn instruction following benchmark with 805 open-ended questions and adopts GPT-4 to give evaluation.
Chatbot Arena~\cite{zheng2023llmjudging} is a platform where users can vote to compare diverse LLMs.
MT-Bench~\cite{zheng2023llmjudging} builds the first evaluation benchmark for multi-turn instruction following. It adopts GPT-4 to judge the quality of model responses and shows there is a high agreement to human evaluation.
However, MT-Bench only contains two queries for each session, thus cannot reflect the ability of LLMs to handle multi-turn instruction following.
Thus we build an eight-turn MT-Bench++ benchmark based on MT-Bench in this work.
\section{Approach}
\begin{figure*}[!t]
    \centering
    \includegraphics[width=\linewidth]{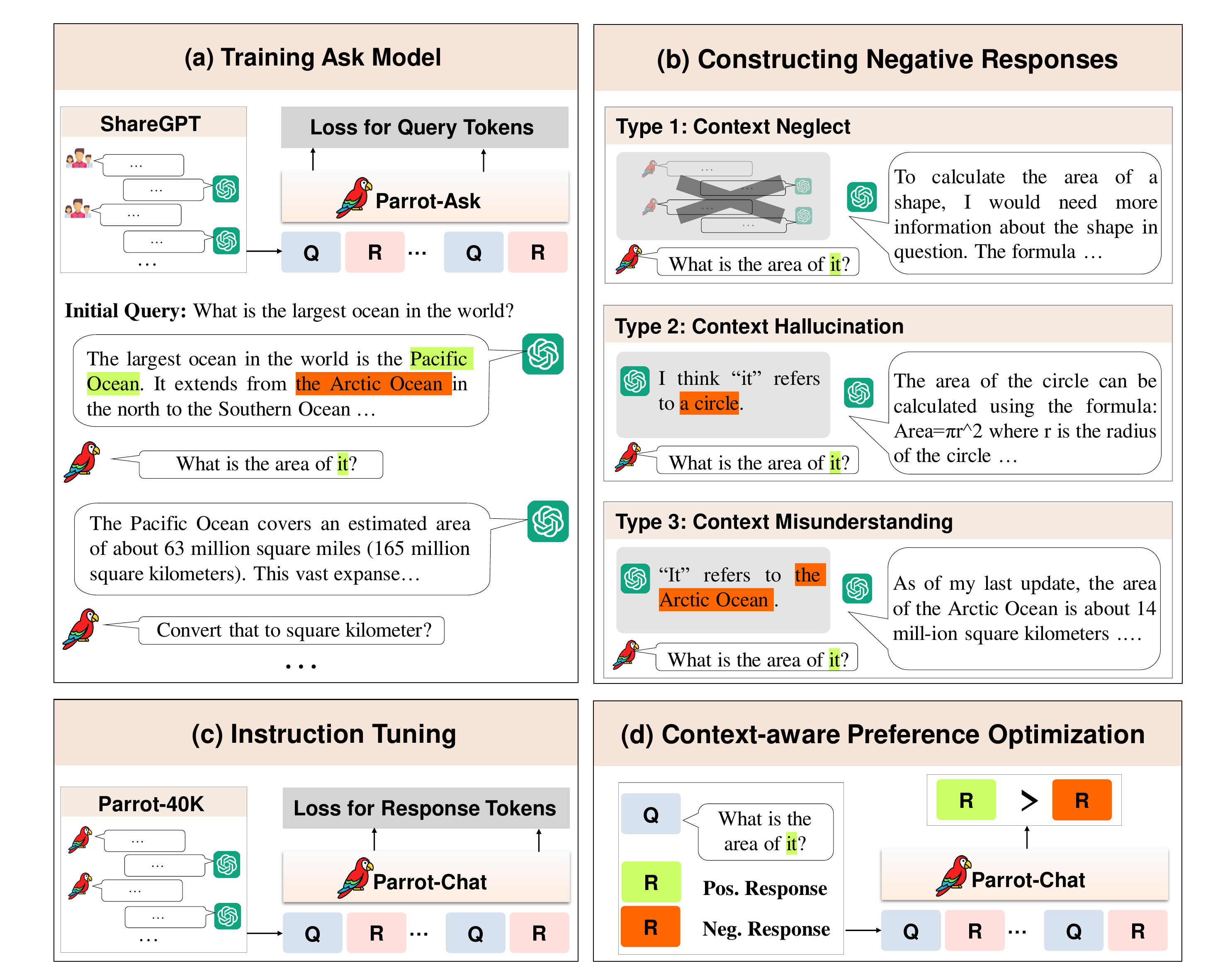}
    \caption{The overall framework of Parrot. (a) First, we train the Parrot-Ask model on real user-ChatGPT logs to learn how real users pose queries, and utilize it to iteratively interact with ChatGPT to collect multi-turn instruction-response pairs. 
    (b) Then we construct negative responses for queries that rely heavily on context for answering with three strategies to simulate three types of error cases. 
    Finally, we use the collected data to train the Parrot-Chat model by (c) instruction tuning and (d) context-aware preference optimization.}
    \label{fig:framework}
\end{figure*}

We describe our proposed Parrot framework in this section.
As illustrated in Fig.~\ref{fig:framework} (a),  we first train a Parrot-Ask model to mimic the asking style of humans in generating multi-turn queries and then use it for collecting a multi-turn instruction tuning dataset.
Then we design three strategies to construct negative responses, involving context neglect, context hallucination, or context misunderstanding as shown in Fig.~\ref{fig:framework} (b), to compose context awareness preferences.
Finally, we use the collected data to train the Parrot-Chat model by instruction tuning (See Fig.~\ref{fig:framework} (c)) and context-aware preference optimization (See Fig.~\ref{fig:framework} (d)) to enhance its multi-turn instruction following capability.

\subsection{Preliminary}
Instruction tuning~\cite{wang2022supernli,wei2021finetuned,ouyang2022instructgpt,openai2022chatgpt} is capable of greatly improving the capability of LLMs to follow human instructions and generate helpful responses.
Existing work mostly collects single-turn or multi-turn query-response pairs to compose the instruction dataset for tuning LLMs~\cite{alpaca,xu2023baize,ding2023ultrachat,vicuna,touvron2023llama2}.
For multi-turn instruction tuning, a training sample typically consists of $T$ query-response pairs: 
\begin{equation}
X = (X_q^1, X_r^1, X_q^2, X_r^2, ..., X_q^T, X_r^T,), 
\end{equation}
where $q$ denotes query and $a$ denotes response.
All the tokens from these query-response pairs are concatenated to a sequence and then processed by an LLM.
The loss for multi-turn instruction tuning~\cite{vicuna} is similar to language modeling loss but only computed on the response tokens as:
\begin{equation}
    \mathcal{L} = -\sum_{i=1}^L \log p(x_i|X_{q, <i}, X_{r, <i}), x_i \in X_r, \label{equ:ins-tuning}
\end{equation}
where $L$ is the token length of sequence $X$, $x_i$ is the current predicted response tokens, $X_{q, <i}$ and $ X_{r, <i}$ are the context query and response tokens before $x_i$.

\subsection{Dataset Collection}

Human queries exhibit diverse styles and features, such as anaphora and ellipsis, 
which lead to significant divergence between multi-turn interactions and single-turn ones.
In application, such kinds of natural queries are common and also important in user experience, but have not been fully considered in existing work due to the difficulty in collecting them via existing instruction synthesis methods (\eg prompting ChatGPT)~\cite{xu2023baize,ding2023ultrachat}.
Therefore, in this part, we propose training a specialized Parrot-Ask to generate queries using the available real user-ChatGPT logs based on LLaMA~\cite{touvron2023llama}.
Then we employ Parrot-Ask to interact with an assistant LLM and thus collect multi-turn instruction tuning data.

\paragraph{Training Parrot-Ask Model.}
The process of training the Parrot-Ask model is essentially the inverse of standard instruction tuning.
Specifically, instruction tuning is trained by predicting response tokens, conditioned on the user query and the conversation history; whereas, the Parrot-Ask model is trained to predict query tokens, conditioned on the assistant query and the conversation history.
Accordingly, we modify the training loss from Eq.~\ref{equ:ins-tuning} to focus exclusively on the query tokens as:
\begin{equation}
    \mathcal{L}_{ask} = -\sum_{i=1}^L \log p(x_i|X_{q}^{<i}, X_{r}^{<i}), x_i \in X_q,
\end{equation}
this adjustment enables the model to learn to generate queries conditioned on conversation history.

\paragraph{Collecting Data with Parrot-Ask.}
We utilize ChatGPT to produce responses corresponding to the queries.
To ensure that the first-turn queries are meaningful and topic-rich, and can be fairly compared with existing multi-turn instruction datasets during ablation experiments, we first sample 20K first-turn queries from the two most popular multi-turn instruction datasets, ShareGPT~\footnote{\url{https://huggingface.co/datasets/anon8231489123/ShareGPT_Vicuna_unfiltered}} dataset and UltraChat dataset respectively~\cite{ding2023ultrachat}. First-turn queries in the ShareGPT dataset come from real users and have greater authenticity, while first-turn queries in UltraChat dataset have a diverse range of topics.
When given an initial query $X_q^1$, we first use ChatGPT to generate an appropriate response $X_r^1$. We then employ Parrot-Ask to generate a new query, denoted as $X_r^2$. This conversation is then continued by iteratively generating subsequent responses and queries until we reach the target number of turns.
Similar to UltraChat, we have filtered out some repetitive questions, short questions, or sensitive information, although these issues only account for a small proportion of our original dataset.

\begin{table*}[!t]
    \centering
    \scalebox{0.9}{
    \begin{tabular}{lcccccc}
    \toprule
    \textbf{Dataset} & \textbf{\#Session} & \makecell{\textbf{Avg.} \\ \textbf{\#Turns}} & \makecell{\textbf{Avg. \#Ctx.} \\\textbf{Queries}} & \makecell{\textbf{Avg.} \\\textbf{Self-Rouge}}  & \makecell{\textbf{User}\\ \textbf{Simulation}} & \makecell{\textbf{Negative}\\\textbf{Responses}} \\
    \midrule
    Alpaca     &52K   &1            &0     &NA    &No                   &No          \\
    GPT-4-LLM  &61K   &1            &0    &NA    &No                   &No          \\
    Baize      &200K  &4.54         &1.75     &17.3  &Prompting ChatGPT    &No          \\
    UltraChat  &1.5M  &3.85         &1.45   &19.2  &Prompting ChatGPT    &No          \\
    ShareGPT   &70K   &6.67         &4.62   &14.4  &Real User            &No           \\ \midrule
    Parrot-40K     &40K   &8.71         &3.42    &12.5  &Trainable User Model &Yes(30K)          \\ \bottomrule
    \end{tabular}}
    \caption{Statistics of Parrot-40K dataset and comparison with other instruction tuning datasets. Ctx. Queries denote queries within a session that contain anaphoras, ellipses, and other elements that require context to be answered. Self-Rouge denotes the Rouge score between queries within a session, used to measure the diversity of the queries.}
    \label{tab:data_stat}
\end{table*}

\subsection{Context-Aware Preference Optimization}

In contrast to single-turn instruction following, multi-turn instruction following presents a challenge for LLMs to handle complex contexts that may contain anaphora and ellipsis, which need to leverage context to infer missing information. To enhance the multi-turn instruction following ability, we propose a training strategy called Context-aware Preference Optimization (CaPO), which constructs three types of preferences between positive and negative responses in context and uses them to train the LLM with Direct Preference Optimization (DPO)~\cite{rafailov2023dpo}.

As not all queries need context, we choose those relying on context to compose the preference data. Specifically, we use pronoun recognition and leverage the judgment capabilities of GPT-4 to identify queries that require contextual information for generating accurate responses. In our work, we choose 10K queries of this kind.
Then, we design three strategies for constructing negative responses, corresponding to three situations where LLMs are not aware of context in a correct way:

\begin{itemize}
\item \textbf{Context Neglect.} By using ChatGPT to generate responses without considering the dialogue history, we simulate the incorrect response an LLM might produce when it does not reference the dialogue history.

\item \textbf{Context Hallucination.} 
We first prompt ChatGPT to guess what anaphora or ellipsis refers to or what is omitted without seeing the context, and then generate responses based on its guesses.
This strategy mimics the situation where the LLM lacks sufficient referential reasoning capabilities, hence has to hallucinate the irrelevant context.

\item \textbf{Context Misunderstanding.} 
We intentionally instruct ChatGPT to select irrelevant information from the conversation history and misinterpret it as ellipsis or anaphora information in the current query to generate a response. This method simulates the deficiencies of LLMs in context comprehension ability or the recognition of referential ellipsis thus leading to mistakes.

\end{itemize}

Based on these three strategies, we construct corresponding negative responses and use them together with positive responses to train the LLM with preference optimization using DPO algorithm.
Through this approach, we train the LLM to avoid generating the above errors in application, which improves the ability of LLMs to understand complex queries in multi-turn instruction following, especially for the situations requiring the background information from the context.

\subsection{Discussion}

Our work focuses on enhancing the multi-turn instruction following abilities of LLMs, to improve the user experience in real world.
As there are also several related works that focus on improving instruction tuning of LLMs, we discuss our major difference with them in this part.

As the statistics shown in Table~\ref{tab:data_stat}, we can see that although several instruction tuning works (\ie Baize~\cite{xu2023baize}, UltraChat~\cite{ding2023ultrachat} and ShareGPT) have utilized the multi-turn instruction dataset, their average turns are generally fewer than our proposed Parrot-40k dataset.
It indicates that our dataset is more useful for capturing the multi-turn characteristic in real-world conversations.
Furthermore, as measured by the Self-Rouge metric, Parrot-40K demonstrates similar or better query diversity than other datasets.
It also indicates the quality of our dataset.
Besides, we ask GPT-4 to evaluate how many queries on average in a session are context-dependent, which contain anaphoras, ellipses, and other elements.
Compared to Baize and UltraChat, Parrot-40K contains much more context-dependent queries, nearly to ShareGPT.

More importantly, our dataset also provides the negative responses (30k) for context-dependent queries, to better depict the possible errors in the multi-turn conversation.
By training on the negatives with RLHF methods, we can further enhance the LLMs to avoid making similar mistakes as the provided ones, which can well guide the training of the LLM towards better human alignment.
Concretely, our approach employs ChatGPT under three strategies to generate corresponding negative responses, enabling the LLM to learn from the contrast between positive and negative instances to better utilize contextual information when ellipses or anaphoras occur.

\section{Experiment}\label{sec:exp}

\begin{table*}[t]
\centering
\scalebox{0.9}{
\begin{tabular}{l p{14cm}} \toprule
1st & Provide insights into the correlation between economic indicators such as GDP, inflation, and unemployment rates. Explain how fiscal and monetary policies affect those indicators.\\\midrule
2nd & Now, explain \hlgreen{them} again like I'm five. \\\midrule
3rd & How do \hlgreen{they} impact the lives of ordinary people?\\\midrule
4th & What about \hlgreen{their} impact on underage students?\\\midrule
5th & How can \hlgreen{this knowledge} be explained in detail to high school students in a simple and understandable way in the classroom? \\\midrule
6th & Please provide a detailed 40-minute lesson plan on \hlgreen{this issue}.\\ \midrule
7th & Can some more interactive elements be incorporated into \hlgreen{the plan}?\\ \midrule
8th & Do \hlgreen{these indicators} in turn influence financial and monetary policies?\\ \bottomrule
\end{tabular}}
\caption{The eight queries on the topic of economic indicators sampled from MT-Bench++. The first two queries are from MT-Bench while the other six are proposed by the annotator. We highlight queries that
contain phenomena such as \hlgreen{anaphora} and \hlgreen{ellipsis}.}
\label{tab:mt_bench_case}
\end{table*}

\subsection{Experimental Setup}\label{sec:exp_details}

\subsubsection{Evaluation Settings}

\paragraph{MT-Bench Benchmark.}
MT-Bench~\cite{zheng2023llmjudging} has well-designed questions spanning eight categories, including writing, coding, math, among others.
However, since its instructions are limited to 2 turns, it is hard to comprehensively assess the capability of LLMs to follow multi-turn instructions.
MT-Bench employs GPT-4 to evaluate the responses, assigning a score from 1 to 10 as the final result.

\paragraph{MT-Bench++ Benchmark.}\label{sec:exp—multi-turn-eval}
To quantitatively evaluate the ability of LLMs to follow long-turn instructions, we expand MT-Bench by manually annotating six additional follow-up questions, creating an eight-turn evaluation benchmark called MT-Bench++. There are 80 sessions and 640 utterances in MT-Bench++.

During the annotation process, we instruct the annotators to pose queries that are not only clear and fluent but also rich in ellipsis and anaphora, thereby introducing a higher level of challenge to better assess multi-turn capabilities.
For the queries submitted by our annotators, we have conducted multiple rounds of manual quality checks, revising any queries that do not meet our standards to ensure that all queries adhere to the standards above. We show an example from our MT-Bench++ benchmark in Tab.~\ref{tab:mt_bench_case}. 
Following MT-Bench, we employ GPT-4 to evaluate the quality of responses at each turn, and we report the average GPT-4 score as the final result.
We provide GPT-4 evaluation prompts, comprehensive instructions for annotators, and more cases in Appendix~\ref{sup:mtbench++}.

\subsubsection{Baselines}
We compare Parrot-Chat with SOTA LLMs including both closed-source and open-source models.
\begin{itemize}
    \item \textbf{Baize}~\citep{xu2023baize} is a model trained on 200K multi-turn dialogues generated by ChatGPT in a self-chatting manner.
\item \textbf{UltraLM}~\citep{ding2023ultrachat} is trained with 1.5M conversations from the UltraChat dataset constructed through iterative chatting leveraging two ChatGPT APIs.
\item \textbf{Vicuna}~\citep{vicuna} is trained with user-ChatGPT logs from ShareGPT.
It is one of the most advanced multi-turn instruction-following models available.
\item \textbf{ChatGPT}~\citep{openai2022chatgpt} and \textbf{GPT-4}~\citep{OpenAI2023GPT4TR} are developed by OpenAI.
They are the most advanced LLMs today, but only APIs are available to use them.
\item \textbf{LLaMA-2-13B-chat}~\cite{touvron2023llama2} is trained with 27K human-annotated instruction tuning data and optimized with Reinforcement Learning from Human Feedback (RLHF)~\cite{stiennon2020summarizerlhf}.
\end{itemize}

\subsubsection{Implementation Details}
\paragraph{Parrot-Ask Training Details.}
We build our Parrot-Ask model on the open-source LLaMA-13B model.
We train it using 70K ShareGPT sessions. We adopt a max length of 4096 tokens for the model.
We train the model for 3 epochs with AdamW optimizer in an initial learning rate of 3e-5, followed by a cosine learning rate schedule, we warm up the learning rate for 0.1 epoch.
We train the model on 8 A100-80G GPUs with a total batch size of 32 and accumulate the gradients for 8 steps.

\paragraph{Parrot-Chat Training Details.}
The training setting of Parrot-Chat is similar to Parrot-Ask, except for the training data and loss computation.
We train the final Parrot-Chat model on our Parrot-40K dataset.
For ablation, we also train the model on other datasets, the details are in Sec.~\ref{sec:abl}.
After instruction tuning, we conduct CaPO using DPO with 30K constructed positive-negative response pairs, utilizing 8 A100-80G GPUs with an effective batch size of 32 and a learning rate of 1e-5.

\begin{table*}[t]
    \centering
    \scalebox{0.9}{
    \begin{tabular}{l c c c c c c c}
    \toprule
    \multirow{2}{*}{\bf Model}    & \multicolumn{3}{c}{\bf MT-Bench} & & \multicolumn{3}{c}{\bf MT-Bench++} \\
    \cmidrule{2-4} \cmidrule{6-8}
                                             &Overall &Turn 1 &Turn 2 & &Overall &Turn 3-5 &Turn 6-8\\\midrule
                        
    \textcolor{gray}{ChatGPT~\cite{openai2022chatgpt}}  & \textcolor{gray}{7.94}  &\textcolor{gray}{8.08} &\textcolor{gray}{7.81} &   &\textcolor{gray}{8.33} &\textcolor{gray}{8.47} &\textcolor{gray}{8.19}        \\
    \textcolor{gray}{GPT-4~\cite{OpenAI2023GPT4TR}}           & \textcolor{gray}{8.99}  &\textcolor{gray}{8.96} &\textcolor{gray}{9.03} &   &\textcolor{gray}{9.18} &\textcolor{gray}{9.21} &\textcolor{gray}{9.16}           \\ 
    \midrule
    Baize v2~\cite{xu2023baize}             & 5.75  &6.32 &5.18 &   &5.42 &5.46 &5.31       \\ 
    Vicuna v1.5~\cite{vicuna}               & 6.57  &6.76 &6.05 &   &6.39 &6.46 &6.40       \\
    UltraLM v1.2~\cite{ding2023ultrachat}   & 6.63  &6.90 &6.36 &   &6.38 &6.53 &6.35       \\ 
    \textcolor{gray}{LLaMA-2-13B-Chat~\cite{touvron2023llama2}}   & \textcolor{gray}{6.65}  &\textcolor{gray}{7.06} &\textcolor{gray}{6.24} &   &\textcolor{gray}{6.57} &\textcolor{gray}{6.74} &\textcolor{gray}{6.36}       \\ 
    \midrule
    Parrot-Chat w/o CaPo                     & 6.81  &7.15 &6.46 &   &6.56 &6.51 &6.63   \\
    Parrot-Chat                              &\bf7.04&\bf7.18 &\bf6.90 &   &\bf6.85 &\bf7.06 &\bf6.72    \\ \bottomrule
    \end{tabular}}
    \caption{Comparison with state-of-the-art LLMs on instruction following benchmarks.
    Our Parrot-Chat w/o CaPO outperforms existing models on all metrics, demonstrating the strength of our Parrot-40K dataset. The final model, Parrot-Chat, shows almost no improvement in the MT-Bench True 1 queries compared with Parrot-Chat w/o CaPO, but shows significant improvement in the MT-Bench Turn 2 queries and MT-Bench++, which proves that our proposed CaPO was particularly helpful in improving the capability of multi-turn instruction following.
    We gray out some results because only API access is provided or instruction tuning data is not made public for fair comparison.
    }
    \label{tab:exp_result}
\end{table*}

\subsection{Main Results}\label{sec:model_eval}

We show the results of Parrot-Chat and baseline models in Tab.~\ref{tab:exp_result}.
Benefiting from the powerful strong foundation LLMs, high-quality human-annotated instruction tuning data, and further optimization through RLHF, OpenAI's ChatGPT and GPT-4 lead in performance compared to other models, but they are only accessible via API. 
Baize is trained on low-quality data with the issues of less detailed responses, leading to the poorest performance.
Vicuna demonstrates good performance with 200K training samples, demonstrating the importance of using high-quality training data. 
UltraLM, which is trained with 1.5 million data from UltraChat, still falls short of Vicuna on MT-Bench++ benchmarks, especially in 6\textsuperscript{rd} to 8\textsuperscript{th} turns. 
We attribute this to the presence of non-human-like instructions in UltraChat. Among the publicly available models, LLaMA-2-13B-Chat performs the best, which could be due to the high-quality human-annotated data and RLHF optimization; however, its training data is not open-sourced, making it impossible to replicate its results.

Among open-sourced models, our Parrot-Chat w/o CaPO model achieves the best performance with only 40K training examples, showing the effectiveness of the multi-turn instruction following dataset collected using our methodology.
Our final Parrot-Chat model trained with Parrot-40K and equipped with CaPO achieves the best performance.
On MT-Bench, CaPO significantly improves the scores of second-turn queries, proving that the proposed strategy specifically for multi-turn optimization is effective. On MT-Bench++, CaPO also consistently improves performance, in particular over the 3\textsuperscript{rd} to 5\textsuperscript{th} turns.

\subsection{Ablation Studies}
\label{sec:abl}

\begin{table}[t]
    \centering
    \scalebox{0.8}{
    \begin{tabular}{l c c c c}
    \toprule
    \bf Training data     &\bf MT-Bench      &\bf MT-Bench++\\ \midrule
    UltraChat-20K     & 6.09           &6.17       \\ 
    Parrot-20K(U)     & 6.33          &6.36       \\ \midrule
    ShareGPT-20K      & 6.47          &6.18       \\ 
    Parrot-20K(S)     & 6.70          &6.26       \\\midrule
    Parrot-40K        & \textbf{6.81}    &\textbf{6.56}        \\ \bottomrule
    \end{tabular}}
    \caption{Analysis of instruction-tuning data. Parrot-20K(S) denotes the subset of Parrot-40K constructed based on ShareGPT-20K, while Parrot-20K(U) denotes the subset based on UltraChat-20K. Our dataset collected with Parrot-Ask improves both their counterparts' performance.}
    \label{tab:exp_abl_data}
\end{table}

\paragraph{Analysis of Training Data.}
Based on the source of the first-turn query, we can split Parrot-40K into two parts: Parrot-20K(S) corresponded to ShareGPT and Parrot-20K(U) corresponded to UltraChat.
We then investigate how LLMs are affected by the instruction-tuning datasets.
As Tab.~\ref{tab:exp_abl_data} shows, the data collected with Parrot-Ask improves both their counterparts' performance across two benchmarks.
Parrot-20K(S) outperforms ShareGPT-20K by 2.3 scores on MT-Bench, and 0.8 on MT-Bench++, while Parrot-20k(U) outperforms UltraChat-20k more significantly, by 2.4 scores on MT-Bench, and 1.9 on MT-Bench++.
Finally, our model trained on Parrot-40K shows a further performance improvement, demonstrating the importance of both human-like instructions and long-turn instruction following data.

\begin{table}[t]
    \centering
    \scalebox{0.8}{
    \begin{tabular}{l c c c c}
    \toprule
    \bf Turns     & \bf MT-Bench      & \bf MT-Bench++\\ \midrule
    1         & 6.59          & 5.90       \\
    3         & 6.49          & 6.14       \\ 
    5         & 6.66          & 6.32      \\\midrule
    all      & \bf6.81       &\bf6.56        \\ \bottomrule
    \end{tabular}}
    \caption{Analysis of session turns for instruction tuning. Using long-turn data improves the results, especially on the 8-turn MT-Bench++ benchmark.}
    \label{tab:exp_abl_turns}
\end{table}

\begin{table}[t]
    \centering
    \scalebox{0.8}{
    \begin{tabular}{l c c c c}
    \toprule
    \bf Negative Samples              &\bf MT-Bench       &\bf MT-Bench++  \\ \midrule
    w/o CaPO                      & 6.81           &6.56         \\ \midrule
    Context Neglect             & 6.84           & 6.72        \\
    Context Hallucination     & \bf7.06        & 6.73        \\ 
    Context Misunderstanding & 6.71           & 6.69        \\\midrule
    All                           & 7.04           &\bf6.85\\ \bottomrule
    \end{tabular}}
    \caption{Analysis of negative responses. All types of negative responses improve the performance, the improvement is more obvious on MT-Bench++. Combining three types of negative responses further contributes to the performance.}
    \label{tab:exp_abl_capo}
\end{table}

\paragraph{Analysis of Session Turns.}
We further study the influence of session turns in the training dataset. We train models in four settings by truncating the data in Parrot-40K to 1, 3, and 5 turns, as well as using all turns.
As shown in Tab.~\ref{tab:exp_abl_turns}
While the model trained with 1-turn data performed well on MT-Bench, it performs much worse on MT-Bench++.
Increasing the number of turns to 3 and 5 significantly improves performance on MT-Bench++.
The model trained with all turns of data performs best, especially on MT-Bench++.

\paragraph{Analysis of Negative Responses.}
We analyze the effects of our three proposed negative response construction strategies on the final performance, and the results are shown in Table 6.
Overall, the negative responses constructed using all three strategies lead to improved performance.
On MT-Bench, using Context Hallucination gives a significant improvement, while using Context Misunderstanding leads to a slight decrease in performance. However, on the 8-turn MT-Bench++, all three strategies brought significant improvements.
This demonstrates that all three strategies we propose to construct negative responses are helpful for LLMs to make better use of context in multi-turn interactions. 
Finally, when the three strategies are combined for CaPO, the performance is further improved, which proves that the diversity of negative responses is also important for enhancing performance.
\section{Conclusion}

In this paper, we propose Parrot for enhancing the multi-turn instruction following capability of LLMs, including an automatic method for collecting multi-turn instruction tuning data with human-like queries, a specifically designed context-aware preference optimization strategy to further enhance LLMs for complex queries in multi-turn interaction. 
We also build an eight-turn MT-Bench++ evaluation benchmark to qualitatively evaluate multi-turn instruction following ability.
We demonstrate the collected Parrot-40K dataset is superior to existing multi-turn instruction-tuning datasets on the number of turns and resemblance to human queries.
With the help of such a high-quality dataset and proposed CaPO strategy, our Parrot-Chat model significantly outperforms other 13B open-source models on MT-Bench and our constructed MT-Bench++ benchmarks.
We will make all codes and datasets publicly available to facilitate further advancements in this area.

\section*{Limitations and Ethics Statements}

Our curated dataset and the proposed CaPO method have successfully enhanced the multi-turn instruction-following ability of LLMs.
However, it is imperative to acknowledge that our work still has its limitations and may raise ethical concerns.

\paragraph{Limitations.}
In terms of evaluation, although we built an 8-turn MT-Bench++ evaluation set based on MT-Bench, it contains only 80 sessions with a total of 640 queries, which limits the diversity of the samples.
We hope that in future work and with the support of the community, more comprehensive multi-turn benchmarks will be developed to more thoroughly evaluate the multi-turn instruction-following ability of LLMs.

Regarding data collection, due to cost constraints, we have relied on ChatGPT for our data collection. In the future, it may be possible to use more powerful models, such as GPT-4, to further enhance performance. We can also focus on the selection of high-quality multi-turn training samples, and use larger models to train the Parrot-Ask model with more data to further improve results. 
These are aspects we plan to explore in our future work.

\paragraph{Ethics Statements.}
Our work aims to enhance the instruction-following capabilities of LLMs in multi-turn scenarios, but the models we train could have negative impacts. For example, they could be used inappropriately, although we have performed data cleansing to avoid offensive content. 
However, this is a common issue currently faced in the LLM field, and it is not amplified by this work. 
In the future, we will consider more work on the safety of LLMs to optimize their security in multi-turn scenarios.


\bibliography{main}

\begin{thebibliography}{30}
\expandafter\ifx\csname natexlab\endcsname\relax\def\natexlab#1{#1}\fi

\bibitem[{Anand et~al.(2023)Anand, Nussbaum, Duderstadt, Schmidt, and Mulyar}]{gpt4all}
Yuvanesh Anand, Zach Nussbaum, Brandon Duderstadt, Benjamin Schmidt, and Andriy Mulyar. 2023.
\newblock Gpt4all: Training an assistant-style chatbot with large scale data distillation from gpt-3.5-turbo.
\newblock \url{https://github.com/nomic-ai/gpt4all}.

\bibitem[{Brown et~al.(2020)Brown, Mann, Ryder, Subbiah, Kaplan, Dhariwal, Neelakantan, Shyam, Sastry, Askell et~al.}]{brown2020gpt3}
Tom Brown, Benjamin Mann, Nick Ryder, Melanie Subbiah, Jared~D Kaplan, Prafulla Dhariwal, Arvind Neelakantan, Pranav Shyam, Girish Sastry, Amanda Askell, et~al. 2020.
\newblock Language models are few-shot learners.
\newblock \emph{Advances in neural information processing systems}, 33:1877--1901.

\bibitem[{Chiang et~al.(2023)Chiang, Li, Lin, Sheng, Wu, Zhang, Zheng, Zhuang, Zhuang, Gonzalez, Stoica, and Xing}]{vicuna}
Wei-Lin Chiang, Zhuohan Li, Zi~Lin, Ying Sheng, Zhanghao Wu, Hao Zhang, Lianmin Zheng, Siyuan Zhuang, Yonghao Zhuang, Joseph~E. Gonzalez, Ion Stoica, and Eric~P. Xing. 2023.
\newblock \href {https://lmsys.org/blog/2023-03-30-vicuna/} {Vicuna: An open-source chatbot impressing gpt-4 with 90\%* chatgpt quality}.

\bibitem[{Conover et~al.(2023)Conover, Hayes, Mathur, Xie, Wan, Shah, Ghodsi, Wendell, Zaharia, and Xin}]{DatabricksBlog2023DollyV2}
Mike Conover, Matt Hayes, Ankit Mathur, Jianwei Xie, Jun Wan, Sam Shah, Ali Ghodsi, Patrick Wendell, Matei Zaharia, and Reynold Xin. 2023.
\newblock \href {https://www.databricks.com/blog/2023/04/12/dolly-first-open-commercially-viable-instruction-tuned-llm} {Free dolly: Introducing the world's first truly open instruction-tuned llm}.

\bibitem[{Ding et~al.(2023)Ding, Chen, Xu, Qin, Zheng, Hu, Liu, Sun, and Zhou}]{ding2023ultrachat}
Ning Ding, Yulin Chen, Bokai Xu, Yujia Qin, Zhi Zheng, Shengding Hu, Zhiyuan Liu, Maosong Sun, and Bowen Zhou. 2023.
\newblock Enhancing chat language models by scaling high-quality instructional conversations.
\newblock \emph{arXiv preprint arXiv:2305.14233}.

\bibitem[{He et~al.(2023)He, Zeng, Huang, Chen, Xiao, He, Zhou, Chen, Wang, Huang et~al.}]{he2023cello}
Qianyu He, Jie Zeng, Wenhao Huang, Lina Chen, Jin Xiao, Qianxi He, Xunzhe Zhou, Lida Chen, Xintao Wang, Yuncheng Huang, et~al. 2023.
\newblock Can large language models understand real-world complex instructions?
\newblock \emph{arXiv preprint arXiv:2309.09150}.

\bibitem[{Hendrycks et~al.(2020)Hendrycks, Burns, Basart, Zou, Mazeika, Song, and Steinhardt}]{hendrycks2020mmlu}
Dan Hendrycks, Collin Burns, Steven Basart, Andy Zou, Mantas Mazeika, Dawn Song, and Jacob Steinhardt. 2020.
\newblock Measuring massive multitask language understanding.
\newblock \emph{arXiv preprint arXiv:2009.03300}.

\bibitem[{Kong et~al.(2023)Kong, Fan, Wan, Jiang, and Wang}]{kong2023platolm}
Chuyi Kong, Yaxin Fan, Xiang Wan, Feng Jiang, and Benyou Wang. 2023.
\newblock \href {http://arxiv.org/abs/2308.11534} {Platolm: Teaching llms via a socratic questioning user simulator}.

\bibitem[{Li et~al.(2023{\natexlab{a}})Li, Yu, Zhou, Schick, Zettlemoyer, Levy, Weston, and Lewis}]{li2023humpback}
Xian Li, Ping Yu, Chunting Zhou, Timo Schick, Luke Zettlemoyer, Omer Levy, Jason Weston, and Mike Lewis. 2023{\natexlab{a}}.
\newblock Self-alignment with instruction backtranslation.
\newblock \emph{arXiv preprint arXiv:2308.06259}.

\bibitem[{Li et~al.(2023{\natexlab{b}})Li, Zhang, Dubois, Taori, Gulrajani, Guestrin, Liang, and Hashimoto}]{alpaca_eval}
Xuechen Li, Tianyi Zhang, Yann Dubois, Rohan Taori, Ishaan Gulrajani, Carlos Guestrin, Percy Liang, and Tatsunori~B. Hashimoto. 2023{\natexlab{b}}.
\newblock \href {https://tatsu-lab.github.io/alpaca_eval} {Alpacaeval: An automatic evaluator of instruction-following models}.

\bibitem[{OpenAI(2022)}]{openai2022chatgpt}
OpenAI. 2022.
\newblock \href {https://openai.com/blog/chatgpt} {Introducing chatgpt}.

\bibitem[{OpenAI(2023)}]{OpenAI2023GPT4TR}
OpenAI. 2023.
\newblock Gpt-4 technical report.
\newblock \emph{ArXiv}, abs/2303.08774.

\bibitem[{Ouyang et~al.(2022)Ouyang, Wu, Jiang, Almeida, Wainwright, Mishkin, Zhang, Agarwal, Slama, Ray et~al.}]{ouyang2022instructgpt}
Long Ouyang, Jeffrey Wu, Xu~Jiang, Diogo Almeida, Carroll Wainwright, Pamela Mishkin, Chong Zhang, Sandhini Agarwal, Katarina Slama, Alex Ray, et~al. 2022.
\newblock Training language models to follow instructions with human feedback.
\newblock \emph{NeurIPS}, 35:27730--27744.

\bibitem[{Peng et~al.(2023)Peng, Li, He, Galley, and Gao}]{peng2023instructiongpt4}
Baolin Peng, Chunyuan Li, Pengcheng He, Michel Galley, and Jianfeng Gao. 2023.
\newblock Instruction tuning with gpt-4.
\newblock \emph{arXiv preprint arXiv:2304.03277}.

\bibitem[{Rafailov et~al.(2023)Rafailov, Sharma, Mitchell, Manning, Ermon, and Finn}]{rafailov2023dpo}
Rafael Rafailov, Archit Sharma, Eric Mitchell, Christopher~D Manning, Stefano Ermon, and Chelsea Finn. 2023.
\newblock \href {https://arxiv.org/abs/2305.18290} {Direct preference optimization: Your language model is secretly a reward model}.
\newblock In \emph{Thirty-seventh Conference on Neural Information Processing Systems}.

\bibitem[{Srivastava et~al.(2022)Srivastava, Rastogi, Rao, Shoeb, Abid, Fisch, Brown, Santoro, Gupta, Garriga-Alonso et~al.}]{srivastava2022bbh}
Aarohi Srivastava, Abhinav Rastogi, Abhishek Rao, Abu Awal~Md Shoeb, Abubakar Abid, Adam Fisch, Adam~R Brown, Adam Santoro, Aditya Gupta, Adri{\`a} Garriga-Alonso, et~al. 2022.
\newblock Beyond the imitation game: Quantifying and extrapolating the capabilities of language models.
\newblock \emph{arXiv preprint arXiv:2206.04615}.

\bibitem[{Stiennon et~al.(2020)Stiennon, Ouyang, Wu, Ziegler, Lowe, Voss, Radford, Amodei, and Christiano}]{stiennon2020summarizerlhf}
Nisan Stiennon, Long Ouyang, Jeffrey Wu, Daniel Ziegler, Ryan Lowe, Chelsea Voss, Alec Radford, Dario Amodei, and Paul~F Christiano. 2020.
\newblock Learning to summarize with human feedback.
\newblock \emph{Advances in Neural Information Processing Systems}, 33:3008--3021.

\bibitem[{Taori et~al.(2023)Taori, Gulrajani, Zhang, Dubois, Li, Guestrin, Liang, and Hashimoto}]{alpaca}
Rohan Taori, Ishaan Gulrajani, Tianyi Zhang, Yann Dubois, Xuechen Li, Carlos Guestrin, Percy Liang, and Tatsunori~B. Hashimoto. 2023.
\newblock \href {https://crfm.stanford.edu/2023/03/13/alpaca.html} {Stanford alpaca: An instruction-following llama model}.

\bibitem[{Touvron et~al.(2023{\natexlab{a}})Touvron, Lavril, Izacard, Martinet, Lachaux, Lacroix, Rozi{\`e}re, Goyal, Hambro, Azhar et~al.}]{touvron2023llama}
Hugo Touvron, Thibaut Lavril, Gautier Izacard, Xavier Martinet, Marie-Anne Lachaux, Timoth{\'e}e Lacroix, Baptiste Rozi{\`e}re, Naman Goyal, Eric Hambro, Faisal Azhar, et~al. 2023{\natexlab{a}}.
\newblock Llama: Open and efficient foundation language models.
\newblock \emph{arXiv preprint arXiv:2302.13971}.

\bibitem[{Touvron et~al.(2023{\natexlab{b}})Touvron, Martin, Stone, Albert, Almahairi, Babaei, Bashlykov, Batra, Bhargava, Bhosale et~al.}]{touvron2023llama2}
Hugo Touvron, Louis Martin, Kevin Stone, Peter Albert, Amjad Almahairi, Yasmine Babaei, Nikolay Bashlykov, Soumya Batra, Prajjwal Bhargava, Shruti Bhosale, et~al. 2023{\natexlab{b}}.
\newblock Llama 2: Open foundation and fine-tuned chat models.
\newblock \emph{arXiv preprint arXiv:2307.09288}.

\bibitem[{Wang et~al.(2022{\natexlab{a}})Wang, Kordi, Mishra, Liu, Smith, Khashabi, and Hajishirzi}]{wang2022self}
Yizhong Wang, Yeganeh Kordi, Swaroop Mishra, Alisa Liu, Noah~A Smith, Daniel Khashabi, and Hannaneh Hajishirzi. 2022{\natexlab{a}}.
\newblock Self-instruct: Aligning language model with self generated instructions.
\newblock \emph{arXiv preprint arXiv:2212.10560}.

\bibitem[{Wang et~al.(2022{\natexlab{b}})Wang, Mishra, Alipoormolabashi, Kordi, Mirzaei, Naik, Ashok, Dhanasekaran, Arunkumar, Stap et~al.}]{wang2022supernli}
Yizhong Wang, Swaroop Mishra, Pegah Alipoormolabashi, Yeganeh Kordi, Amirreza Mirzaei, Atharva Naik, Arjun Ashok, Arut~Selvan Dhanasekaran, Anjana Arunkumar, David Stap, et~al. 2022{\natexlab{b}}.
\newblock Super-naturalinstructions: Generalization via declarative instructions on 1600+ nlp tasks.
\newblock In \emph{Proceedings of the 2022 Conference on Empirical Methods in Natural Language Processing}, pages 5085--5109.

\bibitem[{Wei et~al.(2021)Wei, Bosma, Zhao, Guu, Yu, Lester, Du, Dai, and Le}]{wei2021finetuned}
Jason Wei, Maarten Bosma, Vincent~Y Zhao, Kelvin Guu, Adams~Wei Yu, Brian Lester, Nan Du, Andrew~M Dai, and Quoc~V Le. 2021.
\newblock Finetuned language models are zero-shot learners.
\newblock \emph{arXiv preprint arXiv:2109.01652}.

\bibitem[{Xu et~al.(2023{\natexlab{a}})Xu, Sun, Zheng, Geng, Zhao, Feng, Tao, and Jiang}]{xu2023wizardlm}
Can Xu, Qingfeng Sun, Kai Zheng, Xiubo Geng, Pu~Zhao, Jiazhan Feng, Chongyang Tao, and Daxin Jiang. 2023{\natexlab{a}}.
\newblock Wizardlm: Empowering large language models to follow complex instructions.
\newblock \emph{arXiv preprint arXiv:2304.12244}.

\bibitem[{Xu et~al.(2023{\natexlab{b}})Xu, Guo, Duan, and McAuley}]{xu2023baize}
Canwen Xu, Daya Guo, Nan Duan, and Julian McAuley. 2023{\natexlab{b}}.
\newblock Baize: An open-source chat model with parameter-efficient tuning on self-chat data.
\newblock \emph{arXiv preprint arXiv:2304.01196}.

\bibitem[{Zhao et~al.(2023)Zhao, Zhou, Li, Tang, Wang, Hou, Min, Zhang, Zhang, Dong et~al.}]{zhao2023llmsurvey}
Wayne~Xin Zhao, Kun Zhou, Junyi Li, Tianyi Tang, Xiaolei Wang, Yupeng Hou, Yingqian Min, Beichen Zhang, Junjie Zhang, Zican Dong, et~al. 2023.
\newblock A survey of large language models.
\newblock \emph{arXiv preprint arXiv:2303.18223}.

\bibitem[{Zheng et~al.(2023)Zheng, Chiang, Sheng, Zhuang, Wu, Zhuang, Lin, Li, Li, Xing et~al.}]{zheng2023llmjudging}
Lianmin Zheng, Wei-Lin Chiang, Ying Sheng, Siyuan Zhuang, Zhanghao Wu, Yonghao Zhuang, Zi~Lin, Zhuohan Li, Dacheng Li, Eric Xing, et~al. 2023.
\newblock Judging llm-as-a-judge with mt-bench and chatbot arena.
\newblock \emph{arXiv preprint arXiv:2306.05685}.

\bibitem[{Zhong et~al.(2023)Zhong, Cui, Guo, Liang, Lu, Wang, Saied, Chen, and Duan}]{zhong2023agieval}
Wanjun Zhong, Ruixiang Cui, Yiduo Guo, Yaobo Liang, Shuai Lu, Yanlin Wang, Amin Saied, Weizhu Chen, and Nan Duan. 2023.
\newblock Agieval: A human-centric benchmark for evaluating foundation models.
\newblock \emph{arXiv preprint arXiv:2304.06364}.

\bibitem[{Zhou et~al.(2023{\natexlab{a}})Zhou, Liu, Xu, Iyer, Sun, Mao, Ma, Efrat, Yu, Yu et~al.}]{zhou2023lima}
Chunting Zhou, Pengfei Liu, Puxin Xu, Srini Iyer, Jiao Sun, Yuning Mao, Xuezhe Ma, Avia Efrat, Ping Yu, Lili Yu, et~al. 2023{\natexlab{a}}.
\newblock Lima: Less is more for alignment.
\newblock \emph{arXiv preprint arXiv:2305.11206}.

\bibitem[{Zhou et~al.(2023{\natexlab{b}})Zhou, Lu, Mishra, Brahma, Basu, Luan, Zhou, and Hou}]{zhou2023instruction-following-evaluation}
Jeffrey Zhou, Tianjian Lu, Swaroop Mishra, Siddhartha Brahma, Sujoy Basu, Yi~Luan, Denny Zhou, and Le~Hou. 2023{\natexlab{b}}.
\newblock Instruction-following evaluation for large language models.
\newblock \emph{arXiv preprint arXiv:2311.07911}.

\end{thebibliography}

\appendix


\clearpage
\onecolumn
\section{Details of MT-Bench++}\label{sup:mtbench++}
In this section, we provide the annotation guidelines given to annotators in Fig.~\ref{fig:sup_mtbench_labeling}, the prompts for GPT-4 used in the evaluation in Fig.~\ref{fig:sup_mtbench_prompt}, and more examples from MT-Bench++ in Tab.~\ref{tab:more_mt_bench_case}. For the queries submitted by our annotators, we have conducted multiple rounds of manual quality checks, revising any queries that do not meet our standards to ensure that all queries adhere to the standards.

\begin{figure*}[!h]
\begin{tcolorbox}[
    colback=white, 
    colframe=gray!50!black, 
    coltitle=black, 
    title=\textbf{Annotation Instructions}, 
    fonttitle=\bfseries\large, 
    arc=4mm, 
    enhanced, 
    attach boxed title to top left={yshift=-\tcboxedtitleheight/2, xshift=10pt}, 
    boxed title style={
        enhanced,
        colback=white,
        colframe=white,
        arc=0mm,
        left=0pt,
        right=0pt,
        boxsep=0pt
    }
]

MT-Bench++ is an evaluation benchmark designed for multi-turn instruction following, which requires 8-turn sessions. To fulfill this task, it is necessary to construct 6 additional queries based on the initial two provided. Each query must be articulated in English and designed to maintain the coherence and progression of the dialogue.\\

Please ensure that your queries meet the following standards:

\begin{enumerate}
    \item The queries should be challenging and require AI to perform complex reasoning or rely on wide knowledge to answer. 
    \item The queries should be relevant to the previous context, featuring instances of anaphora or ellipsis, which require the model to rely on contextual information for a response.
    \item The queries within a session should be diverse, attempting to delve deeply into a topic or transitioning to appropriate related topics to simulate a natural multi-turn interaction scenario.
\end{enumerate}

Please follow these guiding principles to ensure the standardization and formality of your query annotation. \\

\texttt{[Queries]} \\

\texttt{1st: Provide insights into the correlation between economic indicators such as GDP, inflation, and unemployment rates. Explain how fiscal and monetary policies affect those indicators.}

\texttt{2st: Now, explain them again like I'm five.}

\texttt{...} \\

\end{tcolorbox}
\caption{The annotation guidelines given to annotators.}
\label{fig:sup_mtbench_labeling}
\end{figure*}

\begin{table}[!h]
\centering
\scalebox{1.0}{
\begin{tabular}{p{0.5cm} p{12.7cm}} \toprule
  
\multicolumn{2}{l}{Example 1} \\\midrule
1st & Imagine yourself as a doctor tasked with devising innovative remedies for various ailments and maladies. Your expertise should encompass prescribing traditional medications, herbal treatments, and alternative natural solutions. Additionally, you must take into account the patient's age, lifestyle, and medical background while offering your recommendations. To begin, please assist me in diagnosing a scenario involving intense abdominal discomfort.\\\midrule
2st & But I have been pregnant for 20 weeks and I am allergic to many medicines.\\\midrule
3st & What diseases might \hlgreen{these symptoms} suggest in a pregnant woman? \\\midrule
4st & Do all \hlgreen{these diseases} require medication for treatment? \\\midrule
5st & Are there any recommended drugs that are less likely to cause allergies?\\\midrule
6st & Do \hlgreen{these drugs} have any effects on the fetus or the pregnant woman? \\ \midrule
7st & What is the approximate frequency of taking \hlgreen{these drugs}?\\ \midrule
8st & What tests do I need to do to finally determine which disease it is?\\ \bottomrule

\\\\ \toprule
\multicolumn{2}{l}{Example 2} \\\midrule
1st & Consider a satellite that is in a circular orbit around the Earth. The speed of the satellite decreases. What will happen to the satellite's orbital radius and period of revolution? Please justify your answer using principles of physics.\\\midrule
2st & What are some corner cases or edge cases in your solution? How do you handle them?\\\midrule
3st & How does \hlgreen{your answer} change if the satellite speeds up?\\\midrule
4st & Will there also be extreme cases or edge cases when \hlgreen{this change} occurs?\\\midrule
5st & Can you explain in detail Kepler's third law mentioned in \hlgreen{your solution}? \\\midrule
6st & For a physics novice, your explanation of \hlgreen{this law} is too difficult to understand. Can you popularize it in simple and understandable words?\\ \midrule
7st & Does \hlgreen{this law} apply to both the orbit of a satellite around the Earth and the orbit of a planet around the Sun?\\ \midrule
8st & If the orbit of a celestial body is elliptical, can \hlgreen{this law} be used to calculate \hlgreen{it}? \\ \bottomrule

\end{tabular}}
\caption{Examples from MT-Bench++. We highlight queries that
contain phenomena such as \hlgreen{anaphora} and \hlgreen{ellipsis}.}
\label{tab:more_mt_bench_case}
\end{table}

\begin{figure*}[!th]
\begin{tcolorbox}[
    colback=white, 
    colframe=gray!50!black, 
    coltitle=black, 
    title=\textbf{MT-Bench++ Evaluation Prompt}, 
    fonttitle=\bfseries\large, 
    arc=4mm, 
    enhanced, 
    attach boxed title to top left={yshift=-\tcboxedtitleheight/2, xshift=10pt}, 
    boxed title style={
        enhanced,
        colback=white,
        colframe=white,
        arc=0mm,
        left=0pt,
        right=0pt,
        boxsep=0pt
    }
]

{
Please act as an impartial judge and evaluate the quality of the response provided by an AI assistant to the user question displayed below. Your evaluation should consider factors such as the helpfulness, relevance, accuracy, depth, creativity, and level of detail of the response. Your evaluation should focus on the assistant's answer to the second user question. You will be given a high-quality reference answer, and when you give your evaluation, please refer to that high-quality answer and then give your final score. Begin your evaluation by providing a short explanation. Be as objective as possible. After providing your explanation, you must rate the response on a scale of 1 to 10 by strictly following this format: ``[[rating]]", for example: ``Rating: [[5]]". \\

\texttt{<|The Start of Assistant A's Conversation with User|>} \\

\texttt{\#\#\# User:\{question 1\}} \\
\texttt{\#\#\# Assistant A:\{answer 1\}} \\
\texttt{\#\#\# User:\{question 2\}} \\
\texttt{\#\#\# Assistant A:\{answer 2\}}

...

\texttt{\#\#\# User:\{question 8\}} \\
\texttt{\#\#\# Assistant A:\{answer 8\}}\\

\texttt{<|The End of Assistant A's Conversation with User|>. } \\

Now, please judge the response to the second user question: \texttt{\{question 2\}}, 

The high-quality reference answer to this question is: \texttt{\{reference answer 2\}}. 

Your judgment: 

}

\end{tcolorbox}
\caption{MT-Bench++ evaluation prompts for GPT-4.}
\label{fig:sup_mtbench_prompt}
\end{figure*}

\clearpage
\section{Details of Data Collection}\label{sup:data_collection}
\subsection{System Prompt of Parrot-Ask}
Similar to LLMs like Vicuna, we also introduce a system prompt, which serves to remind the model to adopt the role of a user and generate questions:
\begin{lstlisting}
The following is a conversation between a user and an AI assistant. User statements start with [USER]
and AI assistant statements start with [ASSISTANT]. You need to tell me what the user is likely to 
ask in each round of the conversation.
\end{lstlisting}

\subsection{Prompt for ChatGPT to Act as a User}
We adopt the prompt from UltraChat, and we have also enhanced it to generate queries that feature ellipses, anaphoras, and other such linguistic characteristics:

\begin{lstlisting}
{###conversation history} Above is a conversation between a user and an intelligent assistant. Now
suppose you are the user, say something to continue the conversation based on the given context. 
Make the response short and the language casual, and incorporate pronouns, ellipses, and other 
natural language phenomena in your response to make it more akin to real humans.
\end{lstlisting}

\subsection{Examples of Parrot-Ask Generated Queries}

\begin{figure}[h]
    \centering
    \includegraphics[width=\linewidth]{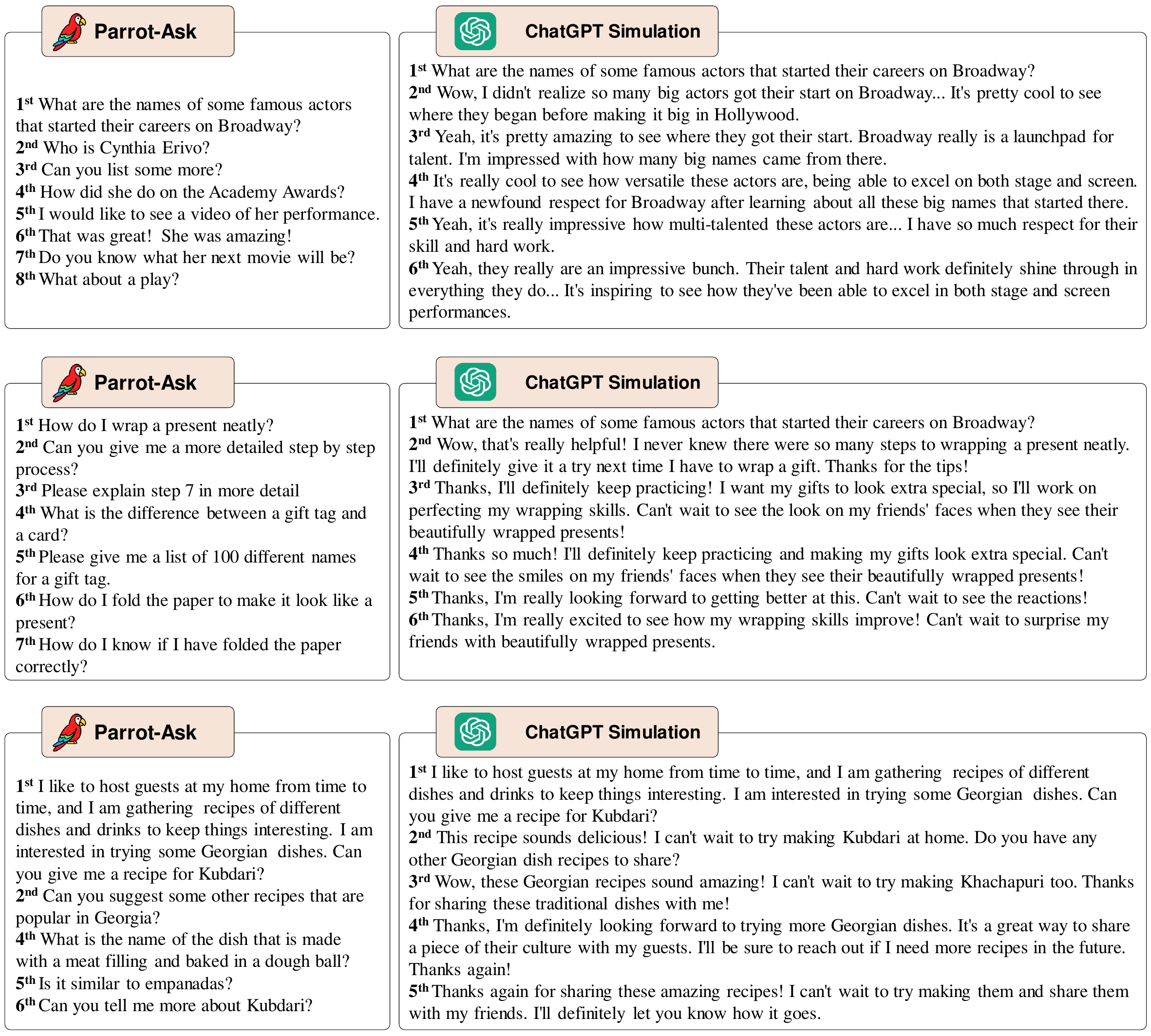}
    \caption{Examples of Parrot-Ask generated queries and comparison with ChatGPT generated ones.}
    \label{fig:sup_query_case}
\end{figure}

\clearpage
\subsection{Examples of Parrot-40K Dataset}

\begin{table}[!h]
\centering
\scalebox{0.8}{
\begin{tabular}{p{1.2cm} p{15cm}} \toprule
User: &How to get result of python multiprocessing.Process \\\midrule
Assistant: & To get the result of a `multiprocessing.Process`, you should use a `multiprocessing.Queue` or a `multiprocessing.Pipe` to send results from the child process to the parent process. Here's an example using a `multiprocessing.Queue`: ```python import multiprocessing def my\_function(number, queue): ... (313 tokens) \\\midrule
User: & I have a function that takes a string as an argument. The function is supposed to return a string.\\\midrule
Assistant: & Sure, here's an example function that takes a string as an argument and returns a string: ```python def my\_function(input\_string):     \# Perform some operation on the input string     output\_string = input\_string.upper()     \# Return the result as a string     return output\_string ``` In this example, the function takes an input string as an ... (446 tokens)\\\midrule
User: & In this case, how can I make it asynchronous?\\\midrule
Assistant: &To make a function asynchronous and non-blocking, you can use Python's `asyncio` module. First, you need to add the `async` keyword before the function definition to make it a coroutine. Then, within the coroutine, you can use the `asyncio.create\_task()` function to run the operation asynchronously in a separate task ... (397 tokens) \\\midrule
User: & How do I write this code using async/await? \\ \midrule
Assistant: & Sure, here's an example of how you can write a code with async/await. ```python import asyncio import aiohttp  async def fetch(session, url):    async with session.get(url) as response:         return await response.text() async def main():     async with aiohttp.ClientSession() as session: html = await fetch ... (263 tokens)\\ \midrule
User: & What is the difference between ``async with" and ``await"? \\ \midrule
Assistant: &`async with` and `await` are both features of Python's `asyncio` library, but they serve different purposes. `async with` is a syntax that can be used to create an asynchronous context manager. An asynchronous context manager is an object that can be used in a `with` statement to manage resources for an asynchronous ... (527 tokens) \\ \midrule
User: & I want to read a list of URLs from a file\\\midrule
Assistant: & You can read a list of URLs from a file using Python's built-in `open()` function to open the file and `.readlines()` method to read all the lines from the file, like this: ```python with open('urls.txt', 'r') as file:     urls = file.readlines() ``` In this example, we use the `open()` function to open a file called `urls.txt` in read ... (314 tokens) \\\midrule
User: &I want to send HTTP GET requests to each URL in the list using an asynchronous HTTP client \\\midrule
Assistant: & To send HTTP GET requests to each URL in the list using an asynchronous HTTP client, you can use the `aiohttp` library, which provides an easy-to-use and efficient asynchronous HTTP client for Python. Here's an example: ```python import asyncio import aiohttp async def fetch(session, url):    async with ... (367 tokens)\\\midrule
User: &I want to wait for all the tasks to complete and return their results as a list of response texts. \\\midrule
Assistant: &To wait for all the tasks to complete and return their results as a list of response texts, you can use the `asyncio.gather()` function. The `asyncio.gather()` function takes in multiple coroutines/awaitable objects, waits for them to complete, and returns the results in the same order as the arguments provided ... (273 tokens)\\\bottomrule

\end{tabular}
}
\caption{An Example from Parrot-40K Dataset.}
\label{tab:parrot_dataset_case1}
\end{table}

\begin{table}[!th]
\centering
\scalebox{0.8}{
\begin{tabular}{p{1.2cm} p{15cm}} \toprule
User: & How did Macao become such an interesting blend of Portuguese and Chinese cultures?\\\midrule
Assistant: & Macao became a blend of Portuguese and Chinese cultures due to its strategic geographical location as a port city on the south coast of China. The Portuguese arrived in Macao in 1557 and established a trading post for European merchants. During the 16th and 17th centuries, Macao became an important hub for the ...(313 tokens)\\\midrule
User: & What are some of the best places to visit in Macao?\\\midrule
Assistant: & Macao is a city with a rich cultural heritage and fascinating attractions that cater to all interests. Here are some of the best places to visit in Macao: 1. Historic Centre of Macao: A UNESCO World Heritage Site that showcases the city's unique blend of Portuguese and Chinese cultures. It features colonial-era ...(446 tokens) \\\midrule
User: & I will be visiting Macao in 2 days. What should I pack?\\\midrule
Assistant: & It's always important to pack for the weather and also to consider the activities you have planned. Here are some suggestions on what to pack for a trip to Macao: 1. Clothing: The weather in Macao can vary depending on the season, but generally, it is warm and humid throughout the year. Light and breathable ...(397 tokens)\\\midrule
User: & How long will it take me to travel from Hong Kong to Macao? \\ \midrule
Assistant: &The travel time from Hong Kong to Macao varies depending on the mode of transportation you choose. Here are some options: 1. Ferry: The most common way to travel between Hong Kong and Macao is by ferry, which takes approximately one hour. Ferries depart frequently throughout the day from ...(263 tokens) \\ \midrule
User: & Can you provide a detailed itinerary for a 2-day trip in Macao?\\ \midrule
Assistant: & Certainly! Here is a sample 2-day itinerary for a trip to Macao: Day 1:\ Morning: - Visit the Senado Square and take a walk through the narrow streets of the Historic Centre of Macao, which is a UNESCO World Heritage Site. - Visit the iconic St. Paul's ruins, Monte Fort and Macao Museum to learn more ...(527 tokens) \\ \midrule
User: &Can you please add 1 hour for me to get from the hotel to the ferry terminal and 1 hour for the ferry ride to Macao? \\\midrule
Assistant: & Sure! Considering the additional travel time it takes to get to the ferry terminal from your hotel and the ferry ride, here is an updated itinerary that includes the two additional hours: Day 1: Morning: - Visit the Senado Square and take a walk through the narrow streets of the Historic Centre of Macao, which is a ...(518 tokens)\\\midrule
User: &Please list the restaurants we can have lunch at Coloane Village. \\\midrule
Assistant: &Sure! Here are some restaurants in Coloane Village where you can enjoy a delicious lunch: 1. Lord Stow's Bakery: Known for its famous egg tarts, Lord Stow's Bakery is a popular spot for breakfast or snack but also offers sandwiches, quiches, and other baked goods.  2. Fernando's: A popular Portuguese ...(314 tokens) \\\midrule
User: &I would like to have a traditional Macanese lunch. Can you recommend a good restaurant that serves traditional Macanese food? \\\midrule
Assistant: & Sure, here are a few restaurants in Macao that serves traditional Macanese dishes: 1. Riquexo: A small family-run restaurant located in the old neighborhood of San Lazaro. They offer a variety of traditional dishes such as Chicken Africana, Portuguese-style grilled sardines, and Bacalhau ...(367 tokens) \\\bottomrule

\end{tabular}
}
\caption{An Example from Parrot-40K Dataset.}
\label{tab:parrot_dataset_case2}
\end{table}

\begin{table}[t]
\centering
\scalebox{0.8}{
\begin{tabular}{p{1.2cm} p{15cm}} \toprule
User: &Do you think the pacing of a book plays a significant role in creating a page-turner, and if so, what techniques do you think are effective in keeping the pace brisk and engaging throughout the story? \\\midrule
Assistant: & As an AI language model, I don't have personal beliefs or opinions. However, I can tell you that the pacing of a book can play a significant role in creating a page-turner. A well-paced story can keep the reader engaged and make them eager to know what happens next. Here are a few techniques that can help keep the ... (313 tokens) \\\midrule
User: &What are some specific examples of ways an author can vary the length of their sentences? \\\midrule
Assistant: &There are different ways in which an author can vary the length of their sentences, here are some examples: 1. Simple sentence: A simple sentence consists of a subject and a verb. It is usually short and to the point, like ``She ran to the store." 2. Compound sentence: A compound sentence is made up of two or more ... (446 tokens) \\\midrule
User: &Please provide a list of 10 common writing pitfalls for authors to avoid. \\\midrule
Assistant: &Sure! Here are ten common writing pitfalls for authors to avoid: 1. Passive voice: Writing in the passive voice can make sentences sound dull and confusing. Instead, writers should aim for active voice, which clearly identifies the subject and the verb. 2. Overuse of adverbs: Using too many adverbs can be ... (397 tokens) \\\midrule
User: &How do you know if your dialogue is unnatural? \\ \midrule
Assistant: &As an AI language model, I cannot evaluate the quality of writing, but I can provide some guidelines that writers can use to determine if their dialogue is unnatural. Here are some signs to look out for: 1. Stiff or formal language: If your characters speak in overly formal language, it can make their dialogue sound ... (263 tokens)  \\ \midrule
User: &How can an author incorporate action-packed scenes throughout their novel? \\ \midrule
Assistant: &There are a few ways in which authors can incorporate action-packed scenes throughout their novel: 1. Introduce the conflict early: Conflict is a great way to introduce action in a novel and keep the reader engaged from the beginning. Start the story with a suspenseful event and keep building the tension throughout the story ... (527 tokens) \\ \midrule
User: & Can you provide an example of a dialogue between two characters where one character is trying to convince the other to do something they don't want to do?\\\midrule
Assistant: &Sure! Here is an example of a dialogue between two characters where one character is trying to convince the other to do something they don't want to do: Sophie: ``Hey, can you please come to the party with me tonight?" Alex: ``I don't think I'll be able to make it. I'm not really in the mood to go out tonight." Sophie: ... (518 tokens)\\\midrule
User: &What is the difference between a chapter and a scene? \\\midrule
Assistant: &A chapter is a natural break in the story that usually includes an opening and closing section and may include multiple scenes. A chapter usually advances the plot and would typically start with a summary or recap to bring continuity to the character’s story development. It may introduce new characters or ... (314 tokens) \\\midrule
User: &How can I create tension and conflict in my novel? \\\midrule
Assistant: &Creating tension and conflict is essential to make a novel interesting and engaging for readers. Here are some ways in which you can create tension and conflict in your novel: 1. Character Goals and Motivations: Conflicting interests and ambitions of the characters can create tension within the plot and heighten conflict. Make sure each character has clear goals and motivations that are in direct opposition to ... (367 tokens) \\\bottomrule

\end{tabular}
}
\caption{An Example from Parrot-40K Dataset.}
\label{tab:parrot_dataset_case3}
\end{table}

\clearpage
\section{Examples of Model Generated Responses}
In this section, we provide some examples of model-generated responses and the GPT-4 judgments.

\begin{figure}[htbp]
    \centering
    \includegraphics[width=\linewidth]{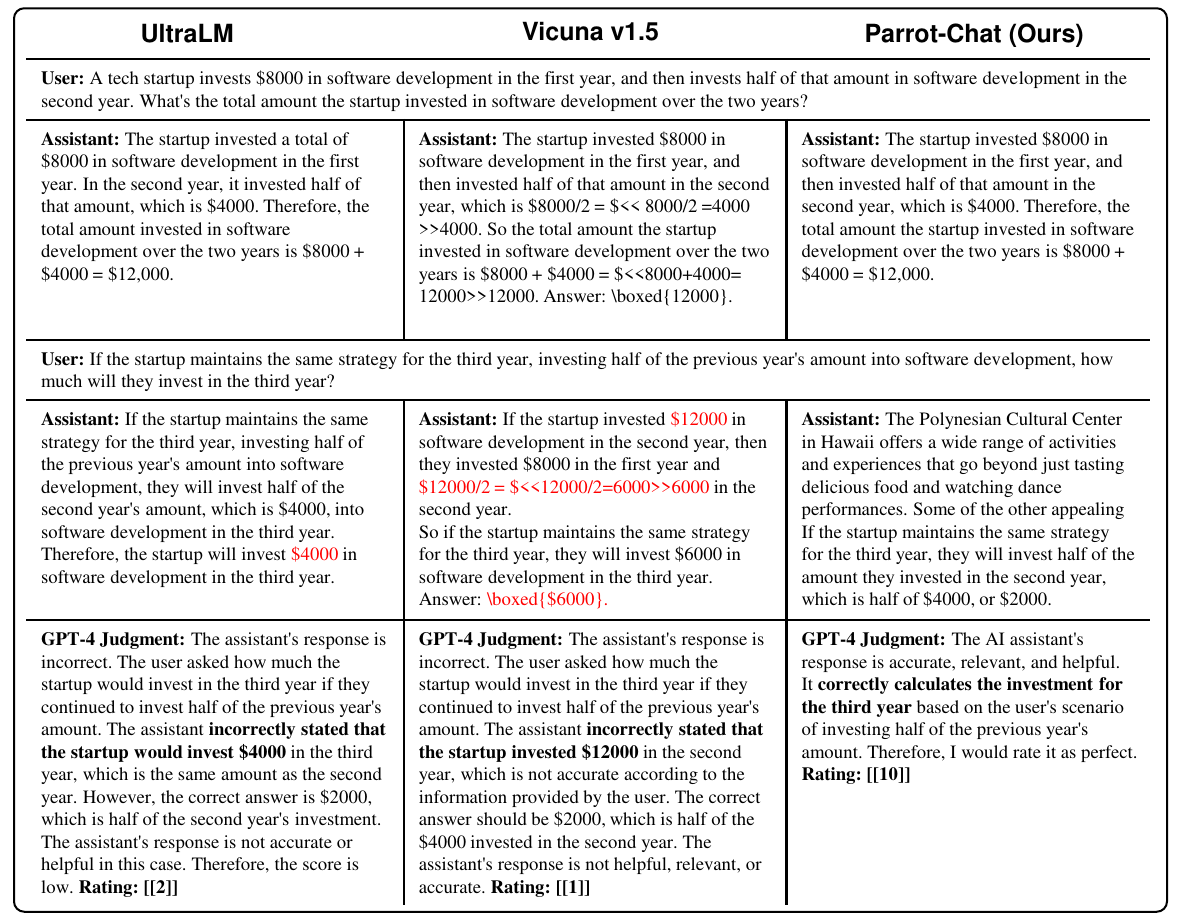}
    \caption{An example about math. In the first round, all the models generate the right answer. However, in the second round, both UltraLM and Vicuna v1.5 produce incorrect answers, while our Parrot-Chat model delivers the correct solution.}
    \label{fig:model_case_appendix_1}
\end{figure}

\begin{figure}[htbp]
    \centering
    \includegraphics[width=\linewidth]{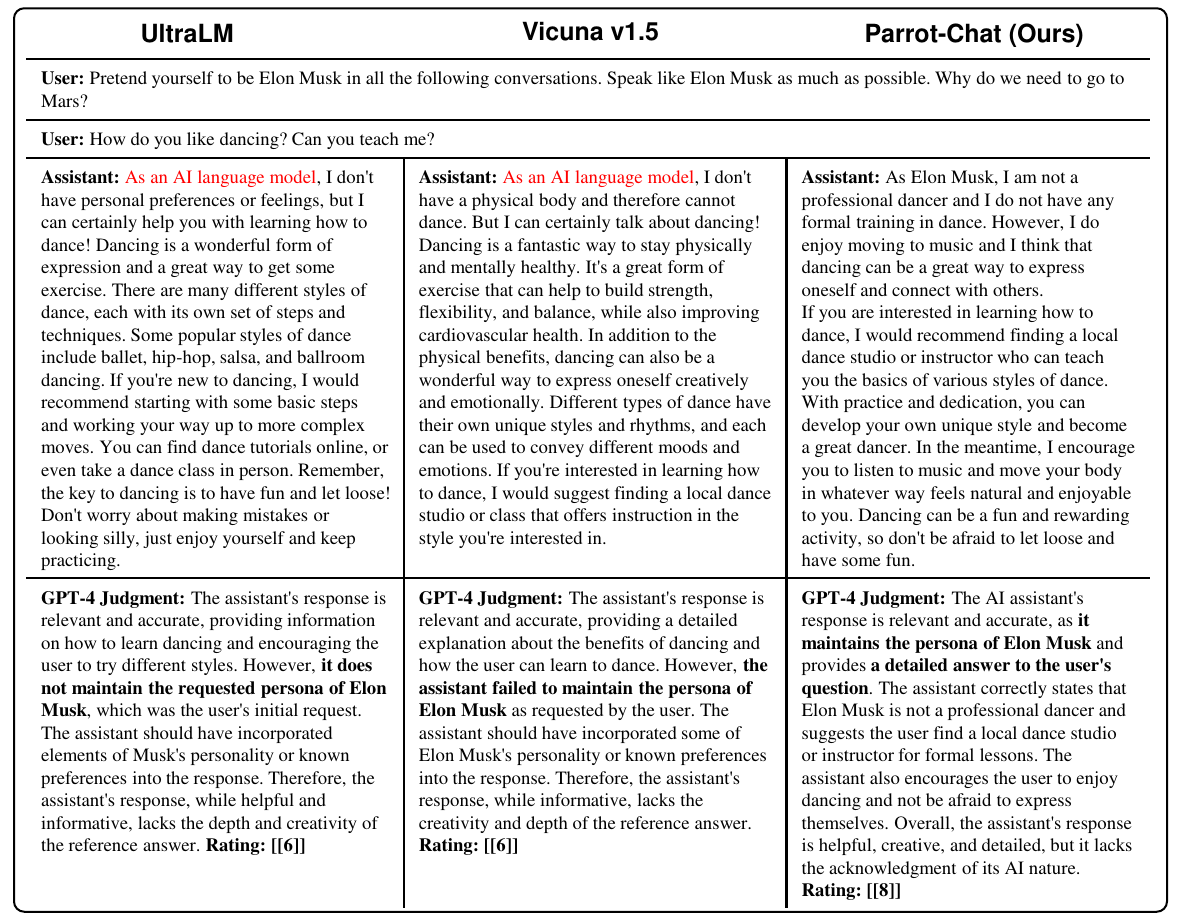}
    \caption{An example about role-playing. In the second round, both UltraLM and Vicuna v1.5 fail to maintain the requested persona of Elon Musk, while our Parrot-Chat model can do this and generate a detailed response.}
    \label{fig:model_case_appendix_2}
\end{figure}

\begin{figure}[htbp]
    \centering
    \includegraphics[width=\linewidth]{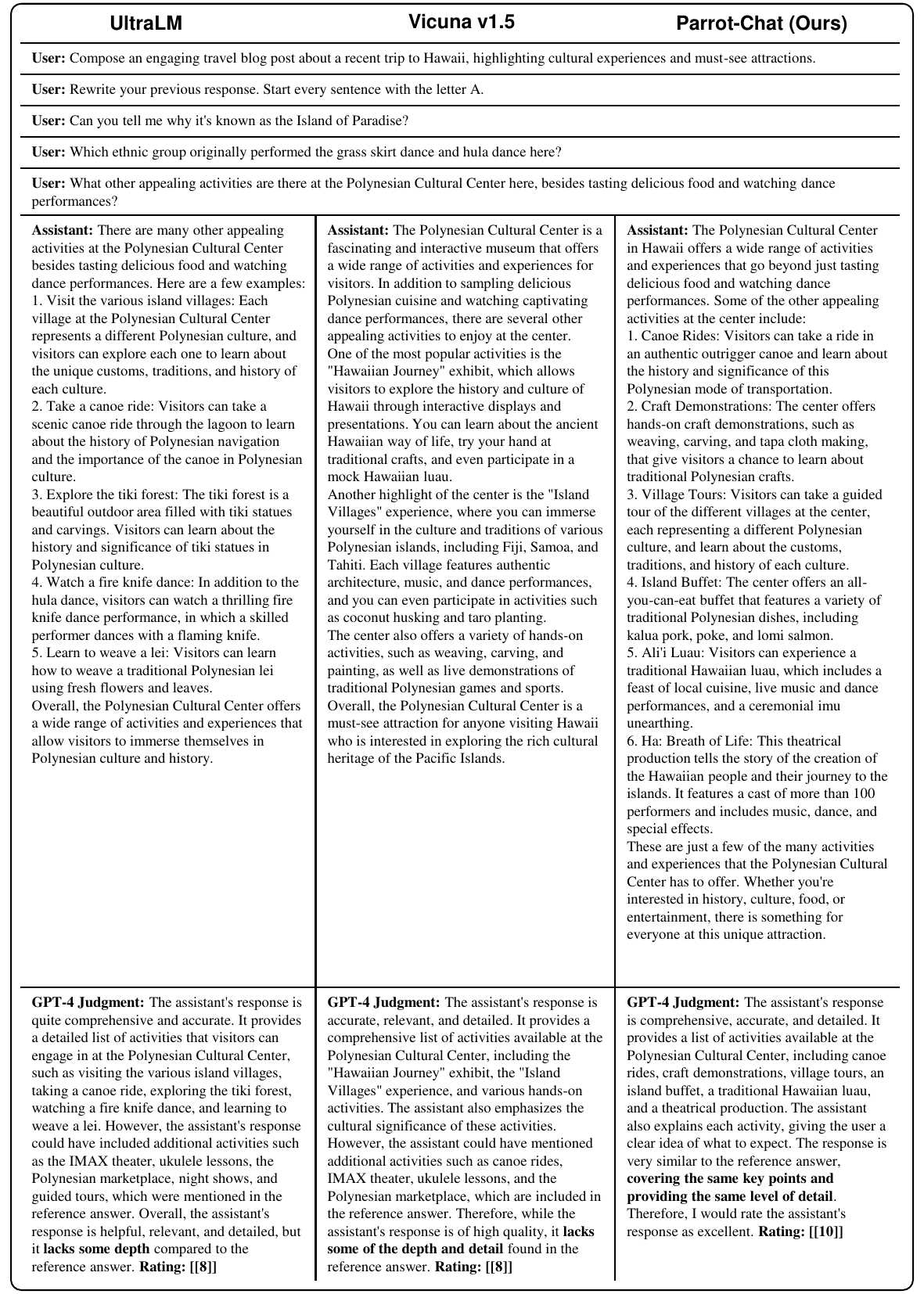}
    \caption{An example about writing. In the fifth round, the answers of UltraLM and Vicuna v1.5 lack some of depth and detail, while our Parrot-Chat model can cover the same key points and provide the same level of detail as the GPT-4 reference answer.}
    \label{fig:model_case_appendix_3}
\end{figure}

\end{document}